\def\year{2019}\relax
\documentclass[letterpaper]{article} 
\usepackage{aaai19}  
\usepackage{times}  
\usepackage{helvet}  
\usepackage{courier}  
\usepackage{url}  
\usepackage{graphicx}  
\usepackage{amsmath}
\usepackage{multirow}
\usepackage{amssymb}

\begin{document}
%
\title{Knowledge Graph Embedding with Entity Neighbors \\ and Deep Memory Network}
\author{Kai Wang \and Yu Liu \and Xiujuan Xu \and Dan Lin\\
  {School of Software Dalian University of Technology}\\
  {Dalian 116620, Liaoning, China} \\
  {\tt kai\_wang@mail.dlut.edu.cn, \{yuliu,xjxu\}@dlut.edu.cn}\\
  {\tt lindan0823@mail.dlut.edu.cn} \\}

\maketitle
\begin{abstract}
Knowledge Graph Embedding (KGE) aims to represent entities and relations of knowledge graph in a low-dimensional continuous vector space. Recent works focus on incorporating structural knowledge with additional information, such as entity descriptions, relation paths and so on. However, common used additional information usually contains plenty of noise, which makes it hard to learn valuable representation. In this paper, we propose a new kind of additional information, called entity neighbors, which contain both semantic and topological features about given entity. We then develop a deep memory network model to encode information from neighbors. Employing a gating mechanism, representations of structure and neighbors are integrated into a joint representation. The experimental results show that our model outperforms existing KGE methods utilizing entity descriptions and achieves state-of-the-art metrics on 4 datasets.
\end{abstract}

\section{Introduction}

With promising potential in artificial intelligence applications, knowledge graphs (KG) have attracted extensive interest \cite{QAKG,KGSSurvey}. Knowledge facts are stored in KG as triplets in form of (head entity, relation, tail entity), e.g. (Apple Inc., Operating Systems Developed, Mac OS). Despite great progress that millions or even billions of facts from real world have been recorded, the construction of large scale knowledge graphs is confronted with incompleteness and sparseness \cite{TransSparse}.

Knowledge graph embedding (KGE) methods have been proposed to overcome this challenge by representing the entities and relations in a low-dimensional continuous vector space \cite{Socher2013}. As one of the typical methods, TransE \cite{TransE} regards every relation as translation between head and tail entities. Benefiting from KGE methods, we can do reasoning and prediction over KG through algebraic computations.

\begin{figure*}
\centering
\includegraphics[width=0.8\textwidth]{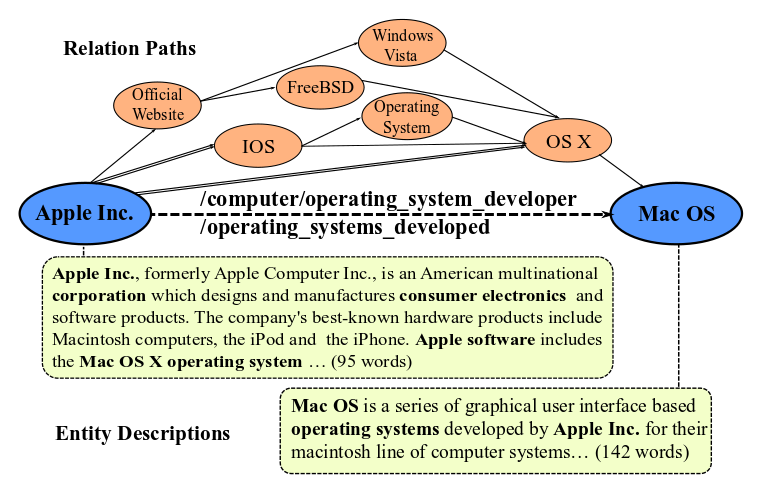}
\caption{Example of entity descriptions and relation paths  for a triplet in Freebase. For relation paths, those paths with less than three intermediate entities are shown in graph. The relation name in paths is ignored, and the double line means two different relations existing between this entity pair.}
\label{fig:1}
\end{figure*}

However, when processing entities with few facts, KGE methods may decline in performance, as they solely learn from fact triplets \cite{Noise}. Therefore, multiple methods have been proposed by incorporating structural knowledge with additional information, including entity descriptions, relation paths and so on. Fig.\ref{fig:1} shows two kinds of additional information of a triplet sampled from Freebase \cite{Freebase}. Apart from fact triplets, those information can provide more semantic or topological features for entity representation.

Although effective, common used additional information usually contains numerous noise, which makes it hard to extract valuable representation. From Fig.\ref{fig:1}, there are 8 different relation paths from `Apple Inc.' to `Mac OS', and all of them pass the `OS X' entity as the last intermediate. It is intuitive that (`Apple Inc.', `OS X', `Mac OS') forms a straight 3-step path, the other longer paths may become redundant. Similarly, many entity descriptions, written by volunteers around the world, are rich in content but not concise. In nearly 100 words long descriptions, key features for an entity are few and far between.

To solve the problem above, we design a new kind of additional information, called entity neighbors. As a man is known by the company he keeps, the representation of an entity can be inferred from its neighbor entities. As an instance, given three words `Corporation', `Mac OS' and `Steve Jobs', people can immediately deduce the `Apple Inc.' entity. Motivated by simplifying entity description and relation paths, we define entity neighbors containing two parts: (1) Semantic Neighbors, mentioned in description of specific entity, or whose description mentions the specific entity; (2) Topological Neighbors, the surroundings of specific entity in KG, having at least one relation with specific entity.

Note that, compared with entity descriptions and neighbor context used in recent methods, the entity neighbors we proposed have three advantages: (1) \textbf{Semantic richness}. Entity neighbors combine both structure features and semantic features. (2) \textbf{Simplicity}. Entity neighbors only retain representative elements while removing a lot of noise. (3) \textbf{Availability}. Entity neighbors can be generated quickly and handle situations where the description text is missing.

In this paper, we propose a novel architecture named Neighborhood Knowledge Graph Embedding (NKGE). We first define entity neighbors consisting of semantic neighbors and topological neighbors. To further generate neighbor representation for each entity, we develop a neighbor encoder based on deep memory network. To the best of our knowledge, it is the first time to utilize memory networks in KG embedding. Based on TransE and ConvE\cite{CONVE} method respectively, we design two kinds of NKGE architecture, combining structure representation and neighbor representation. To verify the effectiveness of entity neighbors and encoder model independently, we design a controlled trial on link prediction task. Experimental results show that our TransE-based model outperforms existing TransE-based methods utilizing the entity descriptions, and  ConvE-based model gets state-of-the-art metrics on most of experimental datasets.

The rest of our paper is structured as follows: We outline related works about KGE and deep memory networks in next section. Section \ref{sec:methods} gives a detailed description of NKGE model, and section \ref{sec:exper} presents experiments to validate the effectiveness of NKGE model. Finally, we summarize this work and the future direction in section \ref{sec:discuss}.

\section{Related Work}
\label{sec:relwork}
\subsection{Knowledge Graph Embedding}
Recent KGE methods can be broadly separated into two groups: translational distance models and semantic matching models \cite{2017Survey}. Represented by the TransE method \cite{TransE}, translational distance models compute the distance between two entities to measure the plausibility of a triplet. To solve flaws in dealing with `1-to-N', `N-to-1', and `N-to-N' relations \cite{TransR}, some variants of TransE such as TransH \cite{TransH}, TransR \cite{TransR} and TransD \cite{TransD} are proposed. Semantic matching models learn representations by matching latent semantics of entities and relations embodied in their vector space representations. RESCAL \cite{RESCAL} models pairwise interactions between latent semantics vectors and latent relation matrix. DistMult \cite{DistMult} simplifies the relation matrices to diagonal matrices. ComplEx \cite{Complex} extends DistMult to better model asymmetric relations. Some of the more recent models achieve strong performances, like ANALOGY\cite{ANALOGY} and ConvE \cite{CONVE}. In our work, we focus on the impact of integrating entity neighbor information in KGE.

\subsection{Incorporating Additional Information}
Common used additional information, as a supplement of structure representation in KG, includes relation paths, entity descriptions and so on. In terms of integrating relation paths, Lin et al.\cite{PTransE} propose path-based TransE (PTransE) to model relation paths, in composition operations of addition, multiplication, and recurrent neural network (RNN). Guu et al.\cite{Guus} utilize KG embedding to answer path queries and built triplets using entity pairs connected with relation paths. Xiong et al.\cite{DeepPath} introduce a reinforcement learning method to learn multi-hop relational paths base on KG embedding. In terms of integrating entity descriptions, Wang et al.\cite{Wang2014} propose a joint model with entity and word embeddings using entity names or Wikipedia anchors. Zhong et al.\cite{Zhong2015} align entity and text representation by entity descriptions in a continuous vector space. Xie et al.\cite{DKRL} jointly learn KG embedding by using CNN model to encode semantics of entity descriptions. Xu et al.\cite{Xu2017} propose a gating mechanism to integrate structural and textual information into a unified architecture. Compared with path-based method, our neighbor processing model are more similar with those joint model using entity descriptions.

\subsection{Deep Memory Network}
Recently, computational models based on attention mechanism and explicit memory have achieved great success in many NLP tasks \cite{Tang2016,MNVD}. Neural Turing Machines \cite{NTM} extend deep neural networks with an external memory, which uses a continuous memory representation with both content and address-based access. Weston et al.\cite{memNN} propose a neural networks based model called Memory Networks, which is designed with non-writable memories, and builds a hierarchical memory representation. Originally designed for question answering tasks, End-to-End Memory Networks (MemN2N) \cite{memN2N} improve Memory Networks to support end-to-end training, and operate via a memory selection mechanism in which relevant memory pieces are adaptively selected based on the input query. Dynamic Memory Networks \cite{DMN} are equipped with an episodic memory and get promising results on question answering and sentiment analysis tasks.

Our model is inspired by the recent success of MemN2N. We aim to utilize the multilayered reasoning capabilities of memory networks for neighbor representation learning.

\section{Methods}
\label{sec:methods}
In this section, we first introduce the notations used in our model. Then, we define two types of entity neighbors and describe the extracting and filtering strategies in detail. After that, we dive into the mathematical and algorithmic details of the deep memory network encoder for neighbor representation. Finally, neighbor representation is utilized in two NKGE architectures, based on TransE and ConvE respectively.

\subsection{Notations}
For a given KG, we define the set of entities as $E$, the set of relations as $R$, and the set of fact triplets as $T$. Each triplet in $T$ is represent by $(h, r, t)$, while $h, t\in E$ and $r \in R$. Each entity $e \in E$ has a set of neighbors $N_e \subseteq E$. Therefore, an entity is represented by two embedding vectors: (1) \textbf{structure embedding}, describes the meaning of entity, and (2) \textbf{neighbor embedding}, constructs neighbor representation. Each relation is represented by a \textbf{relation embedding} vector. All of entity, neighbor and relation embeddings take values in $\mathbb{R}^d$. Our goal is to learn those embedding vectors of all entities and relations.

\subsection{Entity Neighbors}
As a new kind of additional information for KG embedding, entity neighbors refer to a set of entities that are closely related to the specific entity. We define two types of neighbors including topological neighbors and semantic neighbors.

\textbf{Topological neighbors} of an entity are the surrounding of it in given KG. Each neighbor has at least one relation with the specific entity in triplets. Specifically, given an entity $e$, the topological neighbor set of $e$ is $TN(e) = \{t|(e,r,t)\in T, t\in E, r \in R\} \cup \{h|(h,r,e)\in T, h \in E, r \in R\}$. For example, as shown in Figure 1, the topological neighbors of `Apple Inc.' in the graph include `IOS', `OS X' and `Official Website'.

\textbf{Semantic neighbors} of an entity are extracted from description text, including entities mentioned in its description, and entities whose description mentions this entity. Specifically, given an entity $e \in E$, its name is represented by a word set $M_e$ and its description text is a word set $D_e$. The semantic neighbor set of $e$ is $SN (e) = \{ n |\forall n, M_e \subseteq D_n, n \in E, n\neq e\} \cup \{ n |\forall n, M_n \subseteq D_e, n \in E, n\neq e\}$. For example, the semantic neighbors of `Apple Inc.' in Figure 1, include `Corporation', `Mac OS', `OS X', `Apple Software' and so on.

Using the above extracting strategies, we obtain two neighbor sets of an entity. In some cases, there are up to hundreds of neighbors for an entity, such as `the United States'. Therefore, we develop a filter mechanism to select top $K$ typical neighbors from the two sets. First, for each neighbor, we count its number of occurrences in two kinds of neighbor sets respectively. We assume that the lower frequency reflects the neighbor is more representative. Then, given an entity $e$, the neighbors which presenting in both two neighbor sets $( TN(e)\cap SN(e) )$ are selected first. The remaining places are filled by neighbors with smaller frequency in two sets.

Note that, compared with entity descriptions and neighbor context used in recent methods, the entity neighbors we proposed have three advantages: (1) \textbf{Semantic richness}. Entity neighbors combine both structure features and semantic features. (2) \textbf{Simplicity}. Entity neighbors only retain representative elements while removing a lot of noise. (3) \textbf{Availability}. Entity neighbors can be generated from given text quickly and handle situations where the description text is missing.

\subsection{Deep Memory Network Encoder}
After generating entity neighbors, we need to encode the neighbor representation for a given entity. There have been several kinds of neural models used in entity description encoding, such as such as continuous bag-of-words (CBOW), recurrent neural network (RNN) and convolutional neural network (CNN) \cite{DKRL,Xu2017}. However, different from continuous word sequence in description text, each entity neighbor is semantically independent and has potential relations with others. In this paper, we propose a new encoder, DMN encoder based on MemN2N \cite{memN2N}.

\begin{figure}
\centering
\includegraphics[width=0.5\textwidth]{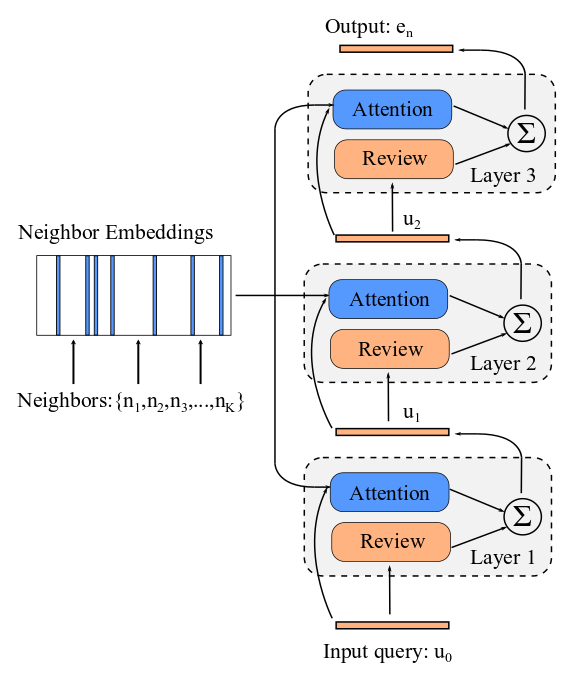}
\caption{An illustration of DMN encoder with three layers.}
\label{fig:2}
\end{figure}

MemN2N is a new RNN-like model, having great performance in question answering tasks. Using sentences as external memory, MemN2N iteratively extracts information by a given query. Attention mechanism is utilized in each iteration to infer potential semantics. Then the final answer is predicted by processing the outputs of the last iteration. Leveraging memN2N's reasoning capabilities, our DMN encoder extracts information from entity neighbors adaptively and integrates it into entity's neighbor representation. The illustration of DMN encoder is shown in Fig. \ref{fig:2}.

The input data of DMN encoder contains input query and external memory. Specifically, given an entity $e$, its neighbor set is $N_e = \{n_1, n_2, \dots n_K\}$, all of them are converted into neighbor embedding vectors. Then $\{\mathbf{n_i}\}$ is used as external memory in DMN encoder, each neighbor $\mathbf{n_i} \in \mathbb{R}^d$ is regarded as a memory cell. The input query $\mathbf{u_0}$ of encoder is a $d$-dimensional vector. Intuitively, an entity should have different neighbor representations under different relations. So given a triplet (h, r, t), we use the relation embedding $\mathbf{r}$ as input query to represent entity $h$ or $t$.

\begin{figure*}
\centering
\includegraphics[width=0.7\textwidth]{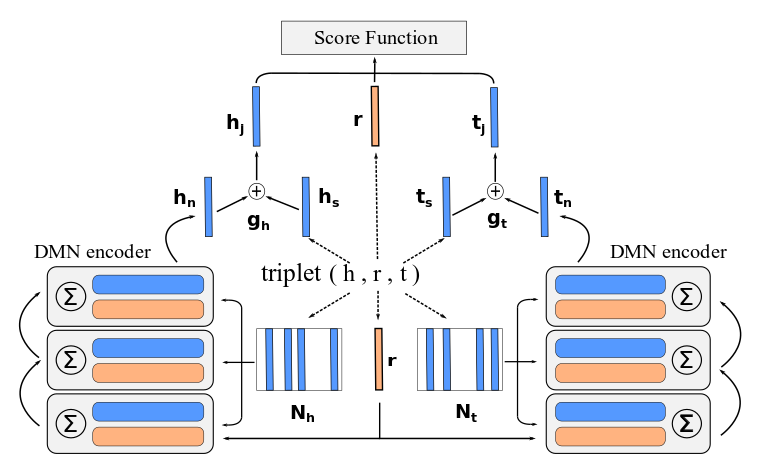}
\caption{The general architecture of NKGE model.}
\label{fig:3}
\end{figure*}

{\bf Single Layer DMN Encoder} We first introduce the DMN encoder with only one layer (iteration), which is made up of two parts: attention part and review part.


In attention part, the attention score for each neighbor $n_i$ is defined as $p_i$, which is
\begin{equation}
g_i = ReLU(\mathbf{W_{att}[n_i;u]}+b_{att})
\end{equation}
\begin{equation}
p_i=softmax(g_i)=\frac{exp(g_i)}{\sum^K_{j=1}exp(g_j)}
\end{equation}
where $\mathbf{W_{att}}\in \mathbb{R}^{1\times 2d}$ and $b_{att}\in \mathbb{R}^{1}$. The $p_i$ score of the $i$th neighbor determines the contribution degree of this neighbor. Then we generate neighbor encoding $\mathbf{o}$ by a weighted sum of neighbors with attention score $\mathbf{p}$:
\begin{equation}
\mathbf{o} = \sum_{i=1}^K p_i \times \mathbf{n_i}
\end{equation}

In review part, we process original information of input query using a fully connected layer independently.
\begin{equation}
\mathbf{\hat{u}} = tanh(\mathbf{W_{rev}u}+b_{rev})
\end{equation}
where $\mathbf{W_{rev}} \in \mathbb{R}^{d \times d}$ and $b_{rev} \in \mathbb{R}^{d}$. Review part's output is added to neighbor encoding $\mathbf{o}$ as the final neighbor representation of single layer DMN encoder.
\begin{equation}
\mathbf{e_n} = \mathbf{o + \hat{u}}
\end{equation}

{\bf Multiple Layer DMN Encoder} Single-layer model of DMN encoder is simple and not powerful enough, while multiple layers allow the deep memory network to learn representation with deep-level abstraction. In the iterative process, the neighbor representation of entity is continuously improved by learning the potential semantics among neighbors.
The multiple layer encoder has several neural layers with the same structure as single-layer model. The input query of the first layer is the same as single-layer encoder, and the output of this layer is used as the input query of upper layer. Finally, the output of the last layer is used as the neighbor representation of $e$.

\subsection{TransE-based NKGE Architecture}
Different from typical KGE techniques, we integrate structure and neighbor information into a joint representation. For a better comparison with recent works \cite{DKRL,Xu2017}, we select the typical method TransE \cite{TransE} to learn structure representation. Given a triplet $(h,t,r)$, TransE's score function is defined as:
\begin{equation}
f((h,t,r)) = \left \| \mathbf{h} + \mathbf{r} -\mathbf{t} \right \|_2^2
\end{equation}
where $\mathbf{h, t} \in \mathbb{R}^d$ are structure embeddings of h, t respectively, and satisfy $\left \| \mathbf{h} \right \| _2^2 = \left \| \mathbf{t} \right \| _2^2 = 1$. $\mathbf{r} \in \mathbb{R}^d$ is relation embedding of r.

To integrate two kinds of entity representations, we use the gating mechanism proposed by \cite{Xu2017}. For each entity $e$, a $d$-dimensional vector $\mathbf{g_e}$ is defined to assign different weight to each dimension in representation vector. To constrain each weight in [0, 1], a logistic sigmoid function is used to transform the gate vector. The joint representation $\mathbf{e_j}$ is computed as follows:
\begin{equation}
\mathbf{e_j}=\sigma {(\mathbf{g_e})} \odot \mathbf{e_s} + (1-\sigma {(\mathbf{g_e})}) \odot \mathbf{e_n}
\end{equation}
where $e_s$, $e_n$ are the structure and neighbor representations of $e$ respectively. Following TransE, our final score function is defined as
\begin{equation}
f_t((h,t,r)) = \left \| \mathbf{h_j} + \mathbf{r} -\mathbf{t_j} \right \|_2^2.
\end{equation}
The general architecture of our model is shown in Fig. \ref{fig:3}.

Following the recent works, we minimize contrastive max-margin criterion \cite{TransE,Socher2013} as objective to train our model. The main idea is that each triplet $(h,r,t)$ from the train set should receive a higher score than a randomly generated triplet. Given a set of fact triplets $T$ as positive sampling set, we generate the negative sampling set $T'$ ($T'\cap T = \phi$):
\begin{equation}
T' = \{(h',r,t)|h'\in E\} \cup \{(h,r,t')|t'\in E\}
\end{equation}
in which each negative sample is derived from a triplet in $T$ by replacing head or tail randomly by another entity. We use the Bernoulli sampling strategy described in \cite{TransH}.
Let the set of all parameters as $\Theta$, we minimize the following objective:
\begin{equation}
\begin{split}
L_t(\Theta ) = \sum_{(h,r,t)\in T}\sum_{(h',r',t')\in T'} max(0,\\
\gamma- f_t(h,r,t) + f_t(h',r',t'))+\eta \left \| \Theta \right \|_2^2
\end{split}
\end{equation}
where $\gamma > 0$ is a margin between positive triplet and negative triplet. We use the standard L2 regularization of all the parameters, weighted by the hyperparameter $\eta$. The optimization is a standard back propagation using stochastic gradient descent (SGD).

\subsection{ConvE-based NKGE Architecture}
ConvE \cite{CONVE} uses a 2D convolutional neural network as the score function and gets start-of-the-art results in several datasets. Given a triplet $(h,t,r)$, ConvE's score function is defined as:
\begin{equation}
f_c((h,t,r)) = f(vec(f([\mathbf{\bar{h}};\mathbf{\bar{r}}]*\mathbf{\omega}))\mathbf{W})\mathbf{t}
\end{equation}
where $\mathbf{\bar{h}}$, $\mathbf{\bar{r}}$ denote a 2D reshaping of $\mathbf{h}$ and $\mathbf{r}$, respectively. The parameters of the convolutional filters the linear layer are denoted as $\mathbf{\omega}$ and $\mathbf{W}$.

Similar with TransE-based architecture, we replace $\mathbf{h}$ in ConvE by the jointly entity representation $\mathbf{h_j}$. Following ConvE's training settings, we apply the logistic sigmoid function $\sigma(\cdot)$ to the scores:
\begin{equation}
p = \sigma(f(vec(f([\mathbf{\bar{h_j}};\mathbf{\bar{r}}]*\mathbf{\omega}))\mathbf{W})\mathbf{t})
\end{equation}
and minimize the following binary cross-entropy loss:
\begin{equation}
L_c(p,y)=-\frac{1}{N}\sum_{i}(y_i \cdot log(p_i)+(1-y_i) \cdot log(1-p_i))
\end{equation}
where the elements of label vector $\mathbf{y}$ are ones for relationships that exists and zero otherwise. Furthermore, we use 1-N scoring, batch normalisation and dropout like ConvE and use Adam \cite{ADAM} as optimiser.

\begin{table}[h]
\setlength{\abovecaptionskip}{10pt}
\setlength{\belowcaptionskip}{0pt}
\centering
\small
\label{tab:1}
\setlength{\tabcolsep}{1mm}{
\begin{tabular}{|c|ccccc|}
\hline
Dataset & \#Rel & \#Ent & \#Train & \#Valid & \#Test \\
\hline
FB15k & $1,345$ & $14,951$ & $483,142$ & $50,000$ & $59,071$ \\
\hline
FB15k237 & $237$ & $14,541$ & $272,115$ & $17,535$ & $20,466$ \\
\hline
WN18 & $18$ & $40,943$ & $141,442$ & $5,000$ & $5,000$ \\
\hline
WN18RR & $11$ & $40,943$ & $86,845$ & $3,034$ & $3,134$ \\
\hline
\end{tabular}}
\caption{Statistics of datasets used in experiments.}
\end{table}

\section{Experiments}
 \label{sec:exper}
 We describe experimental settings and report empirical results in this section.

\begin{table*}
\centering
\small
\label{tab:2}
\begin{tabular}{|c|ccccc|ccccc|}
\hline
\multirow{2}*{Metric} & \multicolumn{5}{c|}{FB15K} & \multicolumn{5}{c|}{FB15K237} \\
~ & MR & MRR & Hits@10 & Hits@3 & Hits@1 & MR & MRR & Hits@10 & Hits@3 & Hits@1\\
\hline
TransE & $111$ & 0.39 & $64.3$ & 47.0 & 24.8 & $221$ & $0.242$ & $42.4$ & 27.6  & 16.6 \\
TransD & $91$ & - & $77.3$ & - & - & - & $0.28$ & $45.3$ & - & - \\
\hline
NKGE(TransE) & $\textbf{50}$ & 0.54 & 78.5 & 62.5 & 39.7 & $\textbf{193}$ & $0.30$ & $48.6$ & 33.4 & 21.7 \\
\hline
DisMult & - & $0.65$ & $82.4$ & 73.3 & 54.6 & $254$ & $0.24$ & $41.9$ & 26.3 & 15.5 \\
ComplEx & - & $0.69$ & $84.0$ & 75.9 & 59.9 & $248$ & $0.24$ & $42.8$ & 27.5 & 15.8 \\
ANALOGY & - & $0.72$ & $85.4$ & 78.5 & 64.6 & - & - & - & - & - \\
ConvE & 51 & $0.66$ & $83.1$ & 72.3 & 55.8 & $244$ & $0.32$ & $50.1$ & 35.6 & 23.7 \\
\hline
NKGE(ConvE) & $56$ & $\textbf{0.73}$ & $\textbf{87.1}$ & \textbf{79.0} & \textbf{65.0} & $237$ & $\textbf{0.33}$ & $\textbf{51.0}$ & $\textbf{36.5}$ & $\textbf{24.1}$ \\

\hline
\end{tabular}
\caption{Results of link prediction task on FB15k and FB15k237}
\end{table*}

\begin{table*}
\centering
\small
\label{tab:3}
\begin{tabular}{|c|ccccc|ccccc|}
\hline
\multirow{2}*{Metric} & \multicolumn{5}{c|}{WN18} & \multicolumn{5}{c|}{WN18RR} \\
~ & MR & MRR & Hits@10 & Hits@3 & Hits@1 & MR & MRR & Hits@10 & Hits@3 & Hits@1\\
\hline
TransE & $251$ & - & $89.2$ & - & - & - & - & 43.2 & -  & - \\
TransD & $212$ & - & $92.2$ & - & - & - & - & 42.8 & - & - \\
\hline
NKGE(TransE) & $\textbf{204}$ & 0.496 & 94.2 & 82.8 & 17.3 & $\textbf{1511}$ & $0.20$ & $50.2$ & 34.4 & 2.2 \\
\hline
DisMult & 901 & $0.822$ & $93.6$ & 91.4 & 72.8 & $5110$ & $0.43$ & $49$ & 44 & 39 \\
ComplEx & - & $0.941$ & $94.7$ & 93.6 & 93.6 & $5261$ & $0.44$ & $51$ & 46 & 41 \\
ANALOGY & - & $0.942$ & $94.7$ & 94.4 & 93.9 & - & - & - & - & - \\
ConvE & 374 & $0.943$ & $93.5$ & 93.5 & 55.8 & $4187$ & $0.43$ & $52.0$ & 44 & 40 \\
\hline
NKGE(ConvE) & $336$ & $\textbf{0.947}$ & $\textbf{95.7}$ & $\textbf{94.9}$ & $\textbf{94.2}$ & $4170$ & $\textbf{0.45}$ & $\textbf{52.6}$ & \textbf{46.5} & \textbf{42.1} \\

\hline
\end{tabular}
\caption{Results of link prediction task on WN18 and WN18RR}
\end{table*}

\subsection{Datasets}
In this paper, we use two popular datasets, FB15k \cite{TransE} and WN18 \cite{WNDS}. FB15k is extracted from Freebase in which a large fraction of content describes knowledge facts about movies, actors, awards, and sports. WN18 is a subset of the English lexical database, WordNet \cite{WordNet}. However, the drawback of two datasets is that many test triplets can be obtained simply by inverting triplets in the training set. To solve this test leakage, FB15k237 \cite{Toutanova2015} and WN18RR \cite{CONVE} are created by removing inverse relations respectively. Statistics of the four datasets are given in Table 1. 
The text descriptions of those datasets are publicly available. We build the entity neighbors data from text description of each entity and triplets in training set for each dataset.

\subsection{Parameter Settings}
For TransE model, we select the margin $\gamma$ among $ \{1.0, 2.0, 4.0\}$, embedding dimension $d$ among $\{ 50, 100, 300\}$, learning rate $\lambda$ among $\{0.001, 0.01, 0.05\}$, the maximum number of neighbors $K$ among $\{10, 20\}$ and the number of layers for multi-layer DMN encoder $L$ among $\{3,6,9\}$ . The dissimilarity measure is set to either L1 or L2 distance. To speed up the convergence and avoid overfitting, the structure embedding of entities and relation embeddings are initialized by pre-trained results of TransE. The neighbor embeddings and rest parameters are initialized by randomly sampling from uniform distribution in $ \left[-0.01, 0.01\right]$. The final optimal configurations are: $\lambda =0.001$, $\gamma=2.0$, $K = 20$, $d = 100$, $L = 6$, $\eta = 1E-5$, and L1 distance. For ConvE model, we set $\lambda =0.0003$, K = 20, d = 200, L = 3, and the rest are the same as the original settings of ConvE.

\subsection{Link Prediction Task}
As a subtask of knowledge graph completion, link prediction aims to predict the missing entity when the other two parts of a triplet $(h, r, t)$ are given. In other word, we need to predict t given $(h, r)$ or predict h given $(r, t)$. Different from other predicting tasks requiring the best one answer, this task focuses on the rank of the correct entity.

We utilize three evaluation metrics similar to \cite{TransE}: (1) Mean Rank (MR), the average rank of all correct entities, (2) Mean Reciprocal Rank (MRR), the average inverse rank for all correct entities, and (3) Hits@N, the proportion of correct entities ranked in top N (N = 1, 3, 10). Lower MR, higher MRR and higher Hits@10 should be achieved by a good embedding model. We also follow the evaluation settings named as `Filter', which removes the candidate triplets appearing in train, valid and test sets before ranking.

\subsection{Results on Four Datasets}
The evaluation results on four datasets are shown in Table 2 and 3. 
We use `NKGE (TransE)', `NKGE (ConvE)' to represent our models based on TransE and ConvE respectively. The baselines are TransE and the state-of-the-art model ConvE. To validate our model's performance, we choose several recent KRE methods, including TransR \cite{TransR}, TransD \cite{TransD}, DisMult \cite{DistMult}, ComplEx \cite{Complex} and ANALOGY \cite{ANALOGY}.
The results show that: (1) Our NKGE models outperform the baselines, TransE and ConvE, on all metrics respectively, which confirms the effectiveness of neighbor representation. (2) NKGE (TransE) gets state-of-the-art Mean Rank on 4 datasets. NKGE (ConvE) gets state-of-the-art MRR and Hits@N across most datasets. (3) On FB15k dataset, the original ConvE is weaker than ComplEx and ANALOGY, while NKGE(ConvE) is very close to ANALOGY, and get better Hits@10.

\subsection{Comparison with entity descriptions}
A controlled trial is designed to compare our model with description-based methods. We choose two representative works, DKRL \cite{DKRL} and Jointly model \cite{Xu2017}, which integrating entity descriptions into structure embeddings. Because both of them based on TransE, we use the TransE-based NKGE model in this trial.

To test the performance of entity neighbors and DMN encoder independently, we design two different derived models: (1) NKGE (CBOW + Nei), using entity neighbors and CBOW encoder, which generate representation by summing up all neighbor embeddings simply; (2) NKGE (Multi + Des), using entity descriptions and multi-layer DMN encoder. The initialization of word embeddings is the same as \cite{Xu2017}. The results are shown in Table 4. 

Compared with two description-based methods on FB15k, our origin model NKGE (DMN + Nei) obtains best scores on MR and Hits@10. NKGE (DMN + Des) outperforms Jointly (A\_LSTM + Des) using the same entity description, which verifies that our DMN encoder has better capability for additional information encoding. In terms of entity neighbors, the derived model NKGE (CBOW + Nei) gets better performance than Jointly (CBOW + Des) using the same CBOW encoder. It proves the validity of entity neighbors as additional information of KG, which can replace entity descriptions to some extent.

\begin{table}[h]
\setlength{\abovecaptionskip}{10pt}
\centering
\small
\label{tab:4}
\begin{tabular}{|l|cc|}
\hline
Metric  & MR & Hits@10\\
\hline
DKRL (CNN + Des) & $91$ &$67.4$ \\
Joinlty (CBOW + Des)  & $92$ &$67.4$ \\
Jointly (A\_LSTM + Des) & $73$ &$75.5$ \\
\hline
NKGE (CBOW + Nei)  & $65$ & $70.0$  \\
NKGE (DMN + Des) & $53$ & $76.0$  \\
NKGE (DMN + Nei) & $\textbf{50}$& $\textbf{78.5}$ \\
\hline
\end{tabular}
\caption{Comparison results of NKGE and description-based methods on FB15k.}
\end{table}

\subsection{Analysis of entity neighbors}
Using two types of neighbors, topological and semantic neighbors, is motivated by simplifying relation paths and entity description. We assume the overlap between two parts is more valuable. To verify this hypothesis, we compare the performance of ConvE-based NKGE using different types of neighbors on FB15k237. As results shown in Table 5, NKGE(T\&S), using the whole entity neighbors, gets better performance than models with only one type of neighbors.

\begin{table}[h]
\setlength{\abovecaptionskip}{10pt}
\centering
\small
\label{tab:5}
\begin{tabular}{|l|cccc|}
\hline
Metric  & MR  & Hits@10 & Hits@3 & Hits@1\\
\hline
NKGE (Top) & $253$ &$50.0$  &$35.7$  &$23.6$\\
NKGE (Sem) & $251$ &$50.2$  &$35.5$  &$23.7$\\
NKGE (T\&S) & $\textbf{237}$ & $\textbf{51.0}$ & $\textbf{36.5}$ & $\textbf{24.1}$ \\
\hline

\end{tabular}
\caption{Comparison results of different types of neighbors on FB15k237.}
\end{table}

\begin{figure}
\centering
\includegraphics[width=0.55\textwidth]{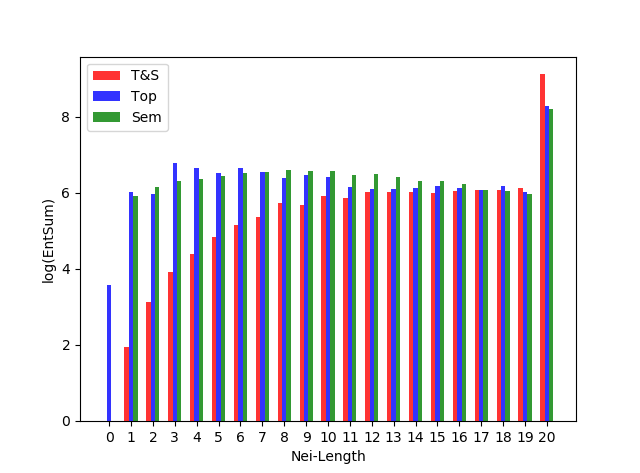}
\caption{The quantitative distribution of different types of neighbors on FB15k237.}
\label{fig:4}
\end{figure}

We also care about the availability of neighbors. As the real KG is usually incomplete and sparse, there are entities with few triplets or missing description. In those case, only using text or paths will get trouble. Containing both two types of neighbors, the entity neighbors we proposed can be effective when one is missing. Fig.\ref{fig:4} shows the quantity distribution of entities with different numbers of neighbors on FB15k237. As the maximum number of neighbors is 20, `T\&S' gets the most entities having complete neighbor information. Note that, there are some entities have no topological neighbors, it reflects the absence situation what we call.

\section{Conclusion}
\label{sec:discuss}
In this paper, we propose the NKGE model for KG embedding. Instead of entity descriptions, we define entity neighbors as new additional information. We explore a deep memory network encoder to extract latent semantics from neighbors. Experiments results show that our model outperforms the baseline TransE and other recent KGE methods on link prediction task. In comparison with description-based methods, both entity neighbors and DMN encoder have better performance.

We will explore the following research directions in future:
\begin{itemize}
\item We select semantic neighbors from text by matching entity name quickly, but inevitably produce omissions. We may design a more effective mechanism in future.
\item The gate mechanism we use only estimates weight according to entity, we may consider relation and neighbor information to improve it.
\item Since entity neighbors are more suitable for sparse KG completion, we will further utilize the NKGE model in real KGs of some specific domains.
\end{itemize}



\bibliography{aaai2019}
\bibliographystyle{aaai}

\end{document}